\documentclass[letterpaper, 10 pt, conference]{ieeeconf}  

\IEEEoverridecommandlockouts                              

\usepackage{graphics}
\usepackage{epsfig}
\usepackage{caption}
\usepackage{subcaption}
\usepackage{color, soul}
\usepackage{diagbox}
\usepackage{amsmath}
\usepackage{amssymb}
\usepackage{hhline}
\usepackage{url}
\usepackage{hyperref}
\newcolumntype{P}[1]{>{\centering\arraybackslash}p{#1}}
\usepackage[T1]{fontenc}

\title{\LARGE \bf Adaptive Energy Regularization for Autonomous Gait Transition and Energy-Efficient Quadruped Locomotion}
\author{
Boyuan Liang\textsuperscript{*}, 
Lingfeng Sun\textsuperscript{*}, 
Xinghao Zhu\textsuperscript{*}, 
Bike Zhang,
Ziyin Xiong,
Yixiao Wang,
Chenran Li, \\
Koushil Sreenath,
Masayoshi Tomizuka%
\thanks{\textsuperscript{*} Equal contributions.}
\thanks{All authors are with the Department of Mechanical Engineering, University of California, Berkeley, California, USA, 94720. }
\thanks{Corresponding author: Lingfeng Sun (email: lingfengsun@berkeley.edu)}
}%

\begin{document}

\maketitle
\thispagestyle{empty}
\pagestyle{empty}


\begin{abstract}
In reinforcement learning for legged robot locomotion, crafting effective reward strategies is crucial. Predefined gait patterns and complex reward systems are widely used to stabilize policy training. Drawing from the natural locomotion behaviors of humans and animals, which adapt their gaits to minimize energy consumption, we investigate the impact of incorporating an energy-efficient reward term that prioritizes distance-averaged energy consumption into the reinforcement learning framework. Our findings demonstrate that this simple addition enables quadruped robots to autonomously select appropriate gaits—such as four-beat walking at lower speeds and trotting at higher speeds—without the need for explicit gait regularizations. Furthermore, we provide a guideline for tuning the weight of this energy-efficient reward, facilitating its application in real-world scenarios. The effectiveness of our approach is validated through simulations and on a real Unitree Go1 robot. This research highlights the potential of energy-centric reward functions to simplify and enhance the learning of adaptive and efficient locomotion in quadruped robots. Videos and more details are at
\href{https://sites.google.com/berkeley.edu/efficient-locomotion}{\url{https://sites.google.com/berkeley.edu/efficient-locomotion}}
\end{abstract}

\section{Introduction}
\label{sec:intro}

Humans and animals exhibit various locomotion behaviors at different speeds, optimizing for their energy efficiency. For instance, humans typically walk at low speeds and run at higher speeds, rarely opting for jumping \cite{cappellini2006motor}. Prior research demonstrated through optimal control on planar models the correlation between speed and optimal gait choices concerning the cost of transport (CoT). For quadrupeds, the optimal gaits were four-beat walking\footnote{Four-beat walking, two-beat walking, trotting and galloping are typical gaits for quadruped robots defined in~\cite{xi2015selectinggaits} based on feet contact schedule.} at low speeds, trotting at intermediate speeds, and trotting/galloping at high speeds~\cite{xi2015selectinggaits}.

Due to the rich information in the gaits, using a gait as guidance for locomotion policies is popular among lots of reinforcement learning (RL) based methods~\cite{siekmann2021simtoreal, margolis2023walk}. However, crafting a versatile and robust locomotion policy that can adapt to and transition between multiple speeds while generalizing across different platforms poses substantial challenges. One of the main challenges here is the reward design. Gait reference can be used as extended state and extra regularization terms in reward functions to provide more supervision or input for low-level MPC~\cite{yang2021fast} controllers.
Previous works~\cite{margolis2023walk, rudin2021learning, feng2023genloco} trained on different quadruped robots within simulation environments like IsaacGym~\cite{liang2018gpu, makoviychuk2021isaac} and successfully transferred these policies to physical hardware. However, they often necessitate intricate reward designs and weight tuning. Apart from gait information, reward terms like feet-air time and contact force penalizing~\cite{rudin2021learning} were also used to encourage specific behaviors and help stabilize the training. 
While these additional reward components are aimed at inducing or preventing specific behavioral traits, they inadvertently align with the broader objective of reducing energy costs \cite{mahankali2024maximizing, fu2021minimizing}. 
This convergence prompts a reconsideration of our reward strategy: Could a more straightforward, energy-centric reward term replace the specifically designed terms used in prior policy training and work for general locomotion tasks, including various linear and angular velocity tracking and terrains? Such a term would encapsulate the core objective of reducing energy consumption and fostering stable and efficient locomotion. 
Gaits at different speeds and terrains are generated posteriorly when the robot finds an energy-efficient way to solve the locomotion task.


\begin{figure}[t]
    \centering
    \includegraphics[width=0.9\linewidth]{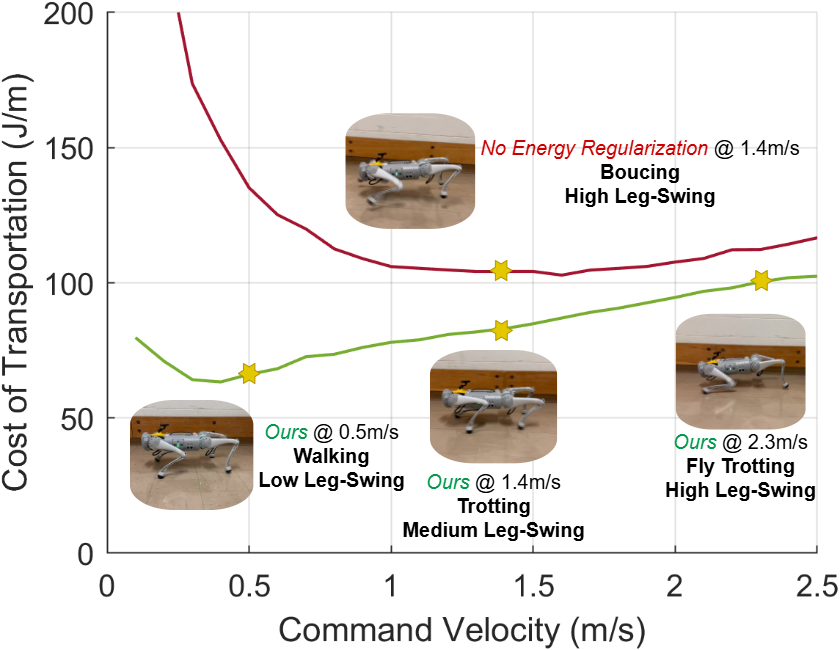}
    \caption{Compared to the baseline when there is no energy regularization, our single policy (from one-time RL training) autonomously adopted different energy-efficient gaits (walking, trotting and fly trotting). It achieved lower energy consumption (adjusted leg-swing) at varying speeds.}
    \vspace{-1em}
    \label{fig:teaser}
\end{figure}


\begin{figure*}[t]
    \centering
    \includegraphics[width=\linewidth]{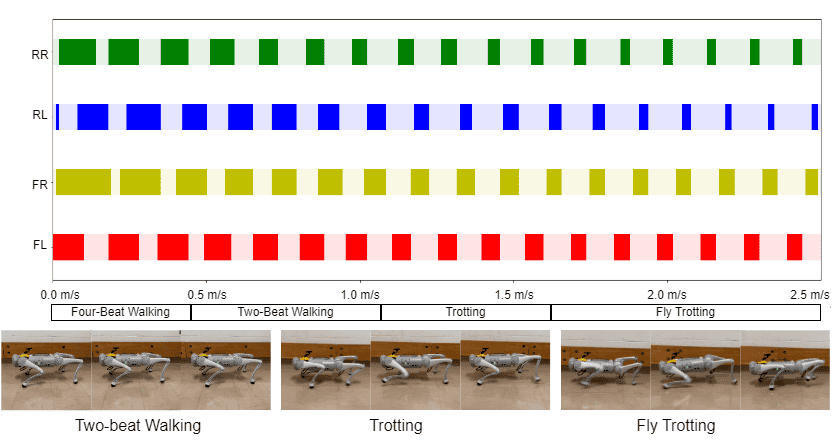}
    \caption{Gait switching under different command velocities. The policy is generated when $\alpha_{en}=1.0$. As the command velocity increases, the policy shows automatic gait transition. We also demonstrate snapshots of two beat walking at 0.5 m/s, trotting at 1.4 m/s and fly trotting at 2.3 m/s.}
    \vspace{-1em}
    \label{fig:LGEL}
\end{figure*}

Building on the concept that energy-efficient gaits correlate with speed \cite{xi2015selectinggaits} and aligning with prior work emphasizing that energy minimization at pre-selected speed results in the emergence of specific gaits \cite{fu2021minimizing}, this study investigates a more streamlined reward formulation for energy-efficient locomotion. By focusing on energy minimization without intricately designed reward components, we aim to verify if such a simplified approach can yield stable and effective velocity-tracking in quadruped robots across various speeds. Instead of generating multiple velocity-specific energy optimal policies~\cite{fu2021minimizing}, we focus on getting a single energy-optimal policy across all target linear and angular velocities, and different terrains via RL training.

In our research, we investigate the impact of incorporating a distance-averaged energy reward term into the reinforcement learning framework. This reward term directly penalizes energy consumption per unit motion traveled, promoting energy-efficient locomotion across various speeds. We explore the effect of different weightings for this energy reward, observing that both excessively low and high weights can lead to undesirable behaviors such as unnatural movements or immobility. By carefully tuning the weight of the distance-averaged energy reward, we demonstrate its effectiveness in facilitating stable velocity tracking and encouraging the emergence of energy-efficient gaits.

Employing this adaptive reward structure within IsaacGym enables the training of robust policies for the Unitree Go1 \cite{go1} quadruped robot. As illustrated in section \ref{sec:exp}, our methodology identifies appropriate gaits, such as four-beat walking at lower speeds and trotting at higher speeds, without predefined gait knowledge. The energy-efficient policy also shows better performance in circle tracking and terrain clearance tasks. The trained single policy is deployed on a real Go1 robot to verify its stable moving and transition locomotion skills in the real world.

The main contribution of this paper includes:
\begin{itemize}
    \item Introduction of a streamlined reward formula integrating basic velocity-tracking and distance-averaged energy minimization to foster stable, velocity-sensitive locomotion policies.
    \item Demonstration that the derived policies autonomously adopt different energy-efficient gaits at varying speeds without preset gait knowledge.
    \item Evaluation of the velocity-tracking and energy efficiency across reward structures and weight tunings for Unitree Go1, culminating in the real-world application of these policies on a Go1 robot, affirming their efficacy in stable locomotion and gait transition.
\end{itemize}

\section{Related Works}
\label{sec:related}

\subsection{Reinforcement Learning for Locomotion Skills}

After deep RL demonstrated its capability to fit general policies in an unsupervised manner, researchers actively sought its potential to be deployed on legged locomotion. Hwangbo et al. \cite{hwangbo2019learning} achieved RL-generated walking policies on a plane ground using a pre-trained actuator net to reduce the sim-to-real gap. Further research also achieved training the policy with adaptive actuator net \cite{ji2022concurrent} or motor control parameters \cite{li2021reinforcement, chen2023learning}. In Hwangbo's follow-up works, locomotion on complicated terrains using a similar approach was also accomplished \cite{miki2022learning, lee2020learning, choi2023learning}. With modern GPU-accelerated simulators \cite{liang2018gpu, makoviychuk2021isaac}, more time-efficient training frameworks were proposed \cite{rudin2021learning, gabriel2024rapid}. Due to the intensive reward engineering in these approaches, the model-free RL tends to converge to a single gait, usually trotting gait, which may not be the most efficient for all terrains and target velocities \cite{raibert1990trotting}.

Many efforts were made to overcome this limitation by promoting the behavioral diversity of legged robots. Researchers in \cite{da2020learning} proposed a hierarchical framework that pre-specifies a set of gait primitives and allows an RL model to choose from them. In \cite{yang2021fast, duan2022sim}, gait primitives were parameterized using the contact schedule and an RL policy was trained to select these parameters. These works used model predictive control (MPC) as the lower-level controller, which demands preliminary knowledge of locomotion and contact modeling \cite{cleach2024fast}. In \cite{margolis2023walk}, behavioral-related arguments such as body height, step frequency, and phase are directly added to the RL model input for end-to-end training. Although various two-beat gaits (where legs touch the ground in pairs) were realized, they cannot generate four-beat gaits (where legs touch the ground in orders) and require manual gait specification in different scenarios. It is more desirable if the quadruped robot can select the most suitable behavior by itself. In \cite{fu2021minimizing}, a pipeline was proposed to output different gaits under different velocities via minimizing energy consumption. However, this pipeline relies on training and distillation of several velocity-specific RL policies.

\subsection{Energy Studies on Locomotion}
The energetic economy of legged robots has always been an important concern for researchers. The cost of transport, namely the amount of energy used per distance traveled, was introduced in~\cite{gabrielli2011price, tucker1975energy}, where the minimization of CoT was connected to the choice of speeds~\cite{ralston1958speed} and step lengths~\cite{umberger2007step} in human locomotion behaviors. The optimal velocity varies for different gaits, allowing animals to transit between different gaits to move at various speeds.

Conceptual legged models of bipedal~\cite{chevallereau2001bipedal, remy2011bipedal}, and quadruped~\cite{kiguchi2002quad, muraro2003quad, xi2015selectinggaits} robots are later designed by researchers to search for energetically optimal motions on various gaits. 
For quadruped robots, genetic algorithms were used in~\cite{kiguchi2002quad} to study potential gaits at different speeds. Our research is mainly inspired by \cite{xi2015selectinggaits}, where optimal control problems are formulated on realistic robot models considering the effects of leg mass, plastic collisions, and damping losses. This work uses an unbiased search on energy-efficient locomotion patterns at various velocities and demonstrates the energy vs. velocity curve for different gaits. Inspired by the results, we believe there exists a policy that can transit between gaits at different velocities with an energy-optimization reward. Among previous works,~\cite{fu2021minimizing} is close to ours in utilizing relations between energy and gaits. It generated multiple policies, with each policy velocity-specific and energy-optimal. In contrast, our work focuses on generating a single energy-optimal policy across various speeds, and it aims to replace the complex-designed reward terms in RL.

\section{Energy Regularization}
\label{sec:method}

A general form of energy regularized locomotion reward takes the following form:
\begin{equation}
    R=(R_{motion}+R_{energy}) * f(R_{aux}),
\end{equation}
where $R_{motion}$ encourages accurate velocity tracking, $R_{energy}$ discourages energy consumption and $R_{aux}$ includes other necessary rewards to stabilize training. $*$ and $f$ are respectively an arithmetic operator and functions. Common choices are $*=+$, $f(x)=-x$ \cite{rudin2021learning} and $*=\times$, $f(x)=\exp(-x)$ \cite{margolis2023walk}.

In previous work \cite{fu2021minimizing}, motion rewards include negative squared linear and angular velocity tracking errors; energy rewards include negative time-averaged motor power with a fixed weight $-\tau\dot{q}$; survival bonus is also added. In experiments, we found such a training process unstable across different speeds, primarily for two reasons—\emph{1)} The negatively unbounded nature of tracking and energy rewards. \emph{2)} The scale of energy reward varies across different speeds, and the energy reward weight usually only works within a narrow range of reference speeds. It is hard to find a single energy reward weight value that works for all reference velocities without knowing more simulation settings and training details. As a result, different speeds are trained in different runs separately in~\cite{fu2021minimizing} to energy-efficient gaits at different speeds.

\begin{figure*}[t]
\centering
    \begin{subfigure}[t]{0.23\textwidth}
        \centering
        \includegraphics[width=\linewidth, trim={5.5cm 10cm 6cm 10.3cm}, clip]{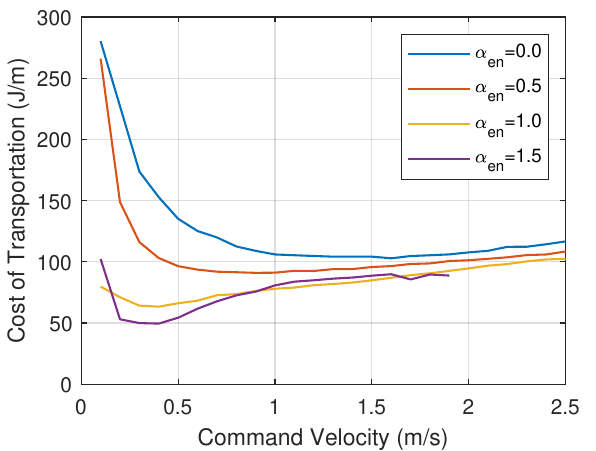}
        \caption{Straight Line Walking}
        \label{subfig:straight_en}
    \end{subfigure}
    \begin{subfigure}[t]{0.23\textwidth}
        \centering
        \includegraphics[width=\linewidth, trim={5.5cm 10cm 6cm 10.3cm}, clip]{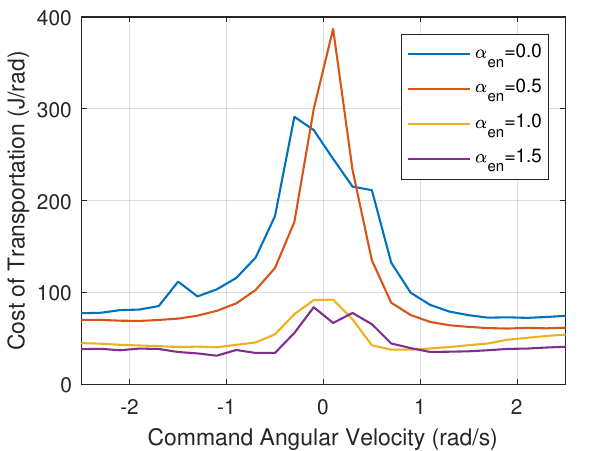}
        \caption{Angular Spining}
        \label{subfig:angular_en}
    \end{subfigure}
    \begin{subfigure}[t]{0.23\textwidth}
        \centering
        \includegraphics[width=\linewidth, trim={5.5cm 10cm 6cm 10.3cm}, clip]{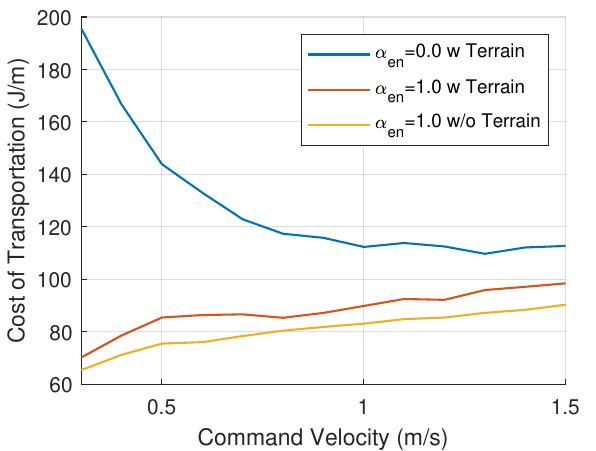}
        \caption{Terrain Walking}
        \label{subfig:terrain_en}
    \end{subfigure}
    \begin{subfigure}[t]{0.23\textwidth}
        \centering
        \includegraphics[width=\linewidth, trim={5.5cm 10cm 6cm 10.3cm}, clip]{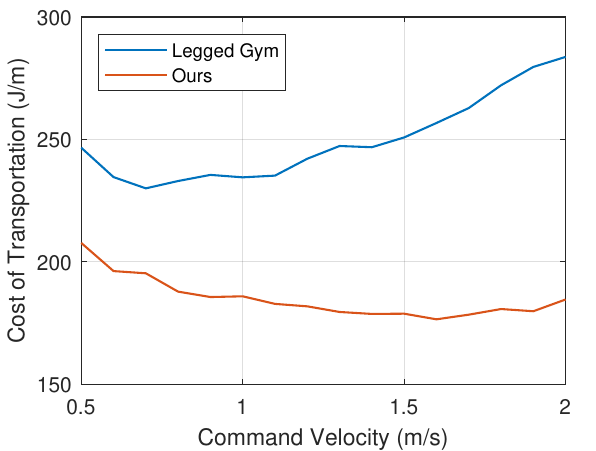}
        \caption{Anymal-C Walking}
        \label{subfig:anymalc_en}
    \end{subfigure}
\caption{Ablation study of energy consumption in Unitree Go1 simulation. For straight line walking, reference linear velocities are chosen from $0.1$ to $2.5$ m/s with $0.1$ common gap. The cost of transportation is measured in $J/m$. For angular spining, reference angular velocities are chosen from $-2.5$ to $2.5$ rad/s with $0.2$ common gap. In both (a) and (b), CoT considerably decreases when $\alpha_{en}$ reaches $1.0$. CoT of $\alpha_{en}=1.5$ when reference velocity of above $1.9$ m/s is not plotted because the output velocity drops to zero in this range. This indicates that velocity tracking accuracy will be sacrificed when energy regularization weight $\alpha_{en}$ is too large. For terrain walking, the robot is asked to walk in a straight line on a rough slope terrain with reference linear velocities from $0.3$ to $1.5$ m/s with $0.1$ common gap, because it is hard to walk either too slowly or too quickly on such terrains. (c) shows that the reduced CoT analogously appears on terrains. (d) shows the effect of energy regularization method to ANYmal-C platform with the same parameters $\sigma_{en,x}$, $\sigma_{en,z}$ and $\alpha_{en}$.}
\label{fig:en_con}
\end{figure*}

To overcome this challenge in reward design, we proposed that the energy-related reward function should depend on the robot's velocity to promote an automatic generation of energy-efficient behavior of legged robots under various reference velocities.
\begin{equation}
    R=(R_{motion}+\alpha_{en}R_{en}(v_x,\omega_z))\exp(-R_{aux})
\label{eq:gen-reward}
\end{equation}
where $v_x$ is the robot moving velocity. The remaining of this section elaborates on each component in (\ref{eq:gen-reward}).

\subsubsection{Motion Rewards} $R_{motion}$ consists of $R_{lin}$ and $R_{ang}$, which respectively encourage the legged robot to track the linear reference velocities in two directions $\hat{v}_x, \hat{v}_y $ and angular reference velocities $\hat{\omega}_z$.
\begin{equation}
\begin{split}
    &R_{motion}=R_{lin}+\alpha_{ang}R_{ang}, \\
    &R_{lin}=\exp\Big(-\frac{|v_x-\hat{v}_x|^2+|v_y-\hat{v}_y|^2}{\sigma_v}\Big), \\
    &R_{ang}=\exp\Big(-\frac{|\omega_z-\hat{\omega}_z|^2}{\sigma_\omega}\Big),
\end{split}
\label{eq:motion-reward}
\end{equation}
where $\hat{v}_y$ and $\hat{\omega}_z$ are not user-specified commands, but randomly sampled during training as explained in section \ref{subsec:exp_setup}. $\sigma_v$ and $\sigma_{\omega}$ are scaling factors depending on the training velocity range.
The structure of motion rewards, the coefficient $\alpha_{ang}=0.5$ for angular velocity tracking, and the scaling coefficients follow the default setting in legged-gym~\cite{rudin2021learning}.

\subsubsection{Energy Rewards} $R_{en}$ rewards the system for consuming less energy while moving.
\begin{equation}
    R_{en}=\exp\Big(-\frac{\sum_{i}|\tau_i||\dot{q}_i|}{\sigma_{en,x}|v_x|+\sigma_{en,z}|\omega_z|}\Big)
\label{eq:en_reward}
\end{equation}

The energy consumption is averaged against the robot's amount of motion. $\tau_i$'s are the actuated torque at each joint while $\dot{q}_i$'s denotes the joint velocities. $\sigma_{en,x}$ and $\sigma_{en,z}$ are energy scaling constants. Components inside the round bracket of (\ref{eq:en_reward}) is equivalent to the linear-distance-averaged energy consumption when $\sigma_{en,z}=0$, which is the conventional definition of the cost of transportation (CoT). Since we are training a linear and angular velocity-dependent policy, we include rotation distance to form a generalized distance while generating the energy reward.

As discussed in \cite{yang2021fast}, directly using the CoT format inside the exponential function in Eq.\ref{eq:en_reward} by detecting it from motion reward fails to generate stable policies via end-to-end training. As a result, authors in \cite{yang2021fast} learn a contact schedule instead and execute it via low-level MPC. In our approach, we deployed the exponential form, which guarantees a positive reward and scale it within $(0,1]$. We multiply the absolute values of each entry of $\tau$ with each entry of $\dot{q}$ and sum them up in (\ref{eq:en_reward}) to follow the fact that a motor does not get charged back even when the applied torque is opposite to the motion~\cite{zhuang2023robot}. This adaptive energy regularization term allows us to learn stable locomotion policies across different speeds, tasks, and embodiments.

One may claim that the selection of an appropriate energy reward weight $\alpha_{en}$ in (\ref{eq:gen-reward}) is still nontrivial since a tiny $\alpha_{en}$ diminishes the influence of energy regularization. At the same time, an overly large value can lead to over-enforcement, compromising velocity tracking accuracy. However, the generalized distance design calibrates the energy reward across different speeds, and the exponential function design on motion and energy reward scale both into $(0,1]$. As a result, a $\alpha_{en}=1$ works for all cases in our experiments, and the parameter is not that sensitive. This trade-off and detailed ablation study will be shown in Section \ref{sec:exp}.

\subsubsection{Auxiliary Rewards}

Energy regularization alone is usually insufficient for generating proper behaviors. To address this, we add a few auxiliary regularization rewards, including collision avoidance, action rate control and trunk orientation regularization. Higher values of these terms signify less desirable performance, so we deployed a negative exponential function as detailed in (\ref{eq:gen-reward}). Comprehensive information on these auxiliary rewards can be found in the \href{https://sites.google.com/berkeley.edu/efficient-locomotion}{project website}.

\section{Experiments}
\label{sec:exp}

The experiments are designed to support the following statements after adding the velocity-dependent energy reward as shown in (\ref{eq:en_reward}) without specifying gait information.
\begin{itemize}
    \item The generated RL policy automatically selects an energy efficient action at different command linear and angular velocity.
    \item The energy reward weight $\alpha_{en}$ should be comparable to motion rewards in order to get a satisfactory RL policy.
    \item The energy reward can also be used for locomotion training on terrains with minor amendments on the auxiliary reward $R_{aux}$.
\end{itemize}

\subsection{Training Setup}
\label{subsec:exp_setup}

We utilized the robot model and PPO training package in \cite{margolis2023walk}. The system outputs the position command of the 12 joints in the next time step. The system inputs include the projected gravity in the robot frame, the commanded x velocities, the commanded yaw rate, each joint's current position and velocity, as well as the action at previous time step. In addition, the inputs of the previous 30 time steps are also given to the RL system. The training episode will reset after 1000 time steps or if any part of the robot except its feet touches the floor.

\subsubsection{Details of Reward Parameters} In motion rewards (\ref{eq:motion-reward}), both $\sigma_v$ and $\sigma_\omega$ were fixed at $0.25$. In energy rewards (\ref{eq:en_reward}), the energy scaling constants are fixed at $\sigma_{en,x}=1000$ and $\sigma_{en,z}=500$. As stated in section \ref{sec:method}, we found that the energy reward $R_{en}$ alone is insufficient to regularize Go1's behavior, which is likely due to the lighter weight compared to its motor power. Thus, following the settings in \cite{margolis2023walk}, we add the fixed auxiliary reward $R_{aux}$. This auxiliary reward is derived mainly from safety concerns, such as penalizing limb-ground collision, out-of-range joint position, and high frequency joint action. The details of $R_{aux}$ can be found on the \href{https://sites.google.com/berkeley.edu/efficient-locomotion}{project website}. Compared to \cite{margolis2023walk}, we did not include any gait-related rewards.

\subsubsection{Curriculum and Domain Randomization} We deployed curriculum technique on command velocities. The sampling range of linear and angular velocities start at $[-1, 1]$ m/s and $[-1, 1]$ rad/s. The sampling range increases when the total reward achieves a certain threshold, and the maximal sampling range is set at $[-5, 5]$ m/s and $[-5, 5]$ rad/s. The ground coefficient of friction is randomized between $[0.05, 1.5]$. We also added a disturbance to the robot mass with a uniform random value in $[-0.1, 3.0]$ kg. A uniformly distributed noise was added to the observation. All these randomized domain parameters are renewed every time the training episode resets, except observation noise is resampled after every time step.

\subsubsection{Amendments for Training on Terrains} All previous techniques are deployed on flat ground trainings. We also tested adding energy regularization while training on terrains. The training terrain shape is rough slope adapted from \cite{rudin2021learning}.

\subsection{Translation and Rotation}


\begin{figure}[t]
    \centering
    \includegraphics[width=\linewidth]{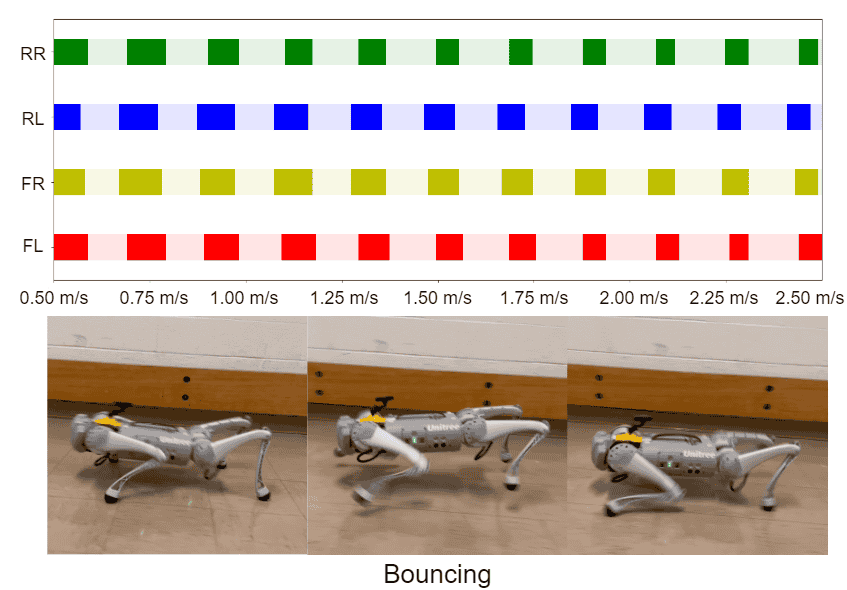}
    \caption{Gait under different command velocities when $\alpha_{en}=0.0$. The policy shows bouncing gait across all command velocities, which is not an energy-efficient choice of gait. We also demonstrate snapshots of bouncing gait, which is the four legs touches and leaves the ground almost simultaneously.}
    \label{fig:LGEL0}
\end{figure}

\begin{figure}[t]
	\centering
    \includegraphics[width=\linewidth, trim={1cm 0cm 1cm 0cm}, clip]{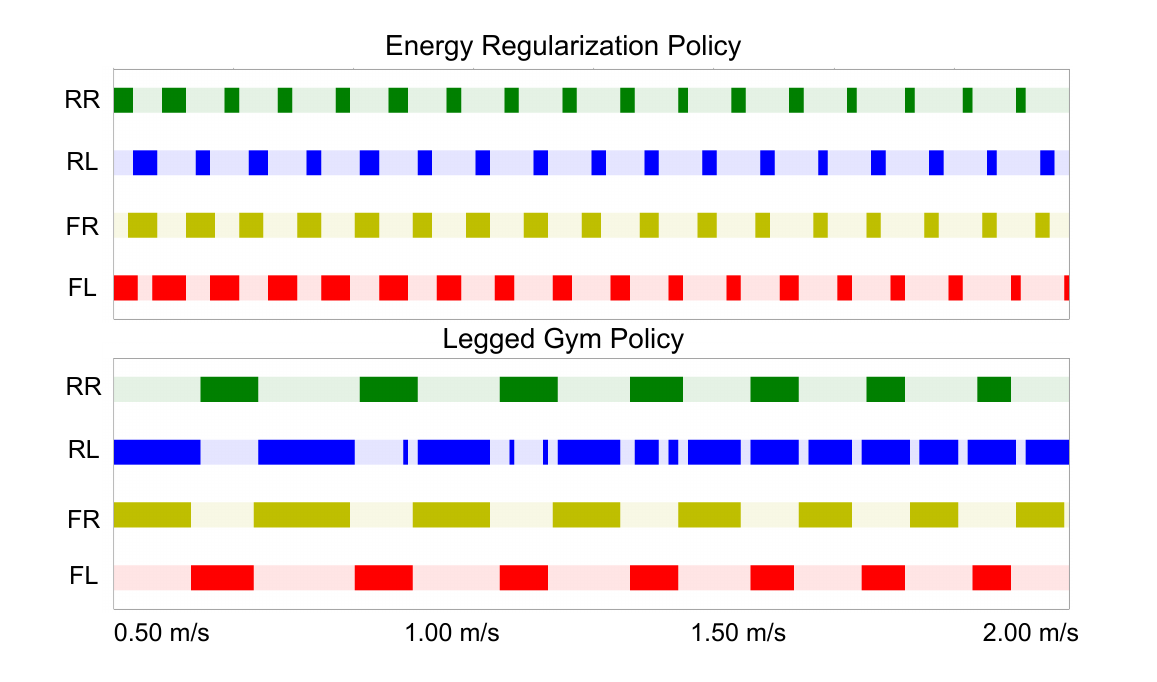}
    \caption{Gait comparison of ANYmal-C between energy regularization and original legged gym policy \cite{rudin2021learning}. Similar gait transition from walking to trotting also appears when energy regularization is applied. In the original legged gym policy, the lifting height of rear right leg is very low, so it has several unexpected mild contacts with the ground. Videos of ANYmal-C simulation can be found on our \href{https://sites.google.com/berkeley.edu/efficient-locomotion}{project website}.}
    \label{fig:anymal-c}
\end{figure}


To demonstrate natural emergence of efficient locomotion gaits, we evaluate the trained walking policy under different reference velocities. Fig. \ref{fig:LGEL} demonstrates a trial run where the legged robot was commanded to move from $\hat{v}_x=0.0$, to $2.5$ m/s. The robot accelerates at $0.5m/s^2$. We plot the gait recorded from the Go1 real-world deployment. We can see that our trained policy exhibits four-beat walking at low speed (around $0.0$ to $0.4$ m/s), where four legs touches the ground in front-left, front-right, rear-left, rear-right sequence. Then, the robot shows two-beat walking (around $0.4$ to $1.1$ m/s) where the four legs touch the ground in diagonal pairs and present observable moments where four legs touches the ground at the same time. At medium speed (around $1.1$ to $1.7$ m/s), the policy exhibits an trotting gait. At this gait, the four legs still touch the ground in diagonal pairs, but there is neither noticeable moment where the four legs touch the ground at the same time, nor moment where all four legs are in the air. At high speed (around $1.7$ m/s and beyond), the trained policy exhibits a fly-trotting gait, which is similar to trotting gait, except there are observable moments where all four legs are in the air. This gait transition is endorsed by previous works \cite{xi2015selectinggaits, yang2021fast} that walking and trotting are respectively the most energy-efficient gaits under low and high speeds.

The result in Fig. \ref{fig:LGEL} is generated with $\alpha_{en}=1.0$. In Fig. \ref{subfig:straight_en} and Fig. \ref{subfig:angular_en}, we compare CoT and linear velocity tracking results across different $\alpha_{en}$. When weight is small (like $0.0$ or $0.5$), the effect of energy regularization is minor, so the generated policy has a much higher CoT. Fig. \ref{fig:LGEL0} shows that when $\alpha_{en}=0.0$, the robot exhibits a bouncing gait across the whole velocity domain, which is not efficient \cite{xi2015selectinggaits}. 

The proposed energy regularization can be similarly applied to other quadruped platforms. Experiments were conducted on the ANYmal-C simulation environment \cite{rudin2021learning} with the same scaling constants $\sigma_{en,x}=1000$ and $\sigma_{en,z}=500$ as well as the regularization weight $\alpha_{en}=1.0$. The generated policy similarly showed preferred gait transition from walking to trotting, which is show in Fig. \ref{fig:anymal-c}. As shown in Fig. \ref{subfig:anymalc_en}, it also successfully reduced CoT in comparison to the original legged gym settings \cite{rudin2021learning}.

\subsection{Circle Tracking}

\begin{figure}[t]
\centering
    \begin{subfigure}[t]{0.5\textwidth}
        \includegraphics[width=1.0\linewidth, trim={5.5cm 11cm 6cm 11.5cm}, clip]{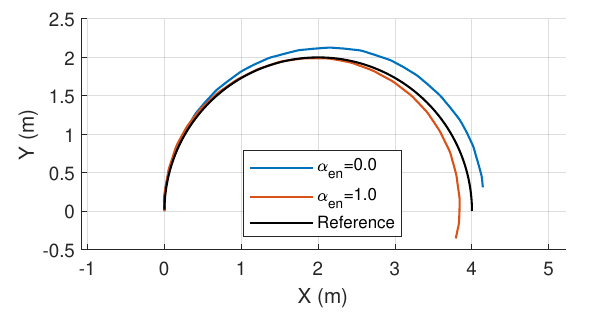}
        \caption{Semicircle Tracking Sketch Plot}
        \label{subfig:STP}
    \end{subfigure}
    \begin{subfigure}[t]{0.5\textwidth}
        \includegraphics[width=1.0\linewidth, trim={5.5cm 11cm 6cm 11.5cm}, clip]{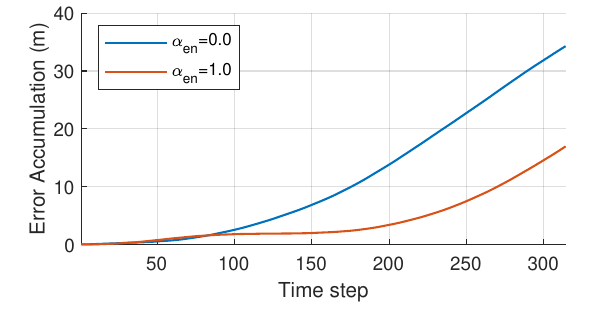}
        \caption{Semicircle Tracking Error Accumulation}
        \label{subfig:STE}
    \end{subfigure}
\caption{Open-loop semicircle tracking comparison between $\alpha_{en}=0.0$ and $\alpha_{en}=1.0$. Since energy regularization reduces random motions irrelevant to target motions, the tracking error for $\alpha_{en}=1.0$ accumulates considerably slower than $\alpha_{en}=0.0$.}
\label{fig:circle}
\end{figure}

We also tested the circle-tracking capacity of the policy. We ask the robot to walk with linear velocity $1.0$ m/s and angular velocity $0.5$ rad/s. This is equivalent to asking the robot to follow a circle with radius $2$ m. We moved the robot in a semicircle with predefined center in simulation and recorded its position at each time step. Assume the distance between the robot and the circle center is $d(t)$ at time step $t$, the error $e(t)$ is defined as the absolute value of the difference between $d$ and the $2$ m pre-defined radius, $|d(t)-2|$. We compare the accumulated error $\sum_{i=1}^Te(t)$ between $\alpha_{en}=0.0$ and $1.0$ policies.

The results are demonstrated in Fig. \ref{fig:circle}. The trajectory radius of $\alpha_{en}=1.0$ policy is closer to the $2$ m target and the error also accumulates much slower. This is because that energy regularization can effectively reduce motions that are not directly related to the target $1.0$ m/s, $0.5$ rad/s motion.

\subsection{Terrain Clearance}

The effect of energy regularization on quadruped gait can also be deployed on terrain clearance. Since it is hard to walk either very fast or very slow on a rough slope, the training velocity range is limited to only $[-1.5, 1.5]$ m/s. When no energy regularization presents, the quadruped robot also tends to show a bouncing gait. When $\alpha_{en}=1.0$, a more natural trotting gait appears. As indicated in Fig. \ref{fig:en_con}, the policy generated from $\alpha_{en}=1.0$ is also a more energy efficient gait.

\begin{figure}
\centering
    \begin{subfigure}{\linewidth}
        \includegraphics[width=\linewidth]{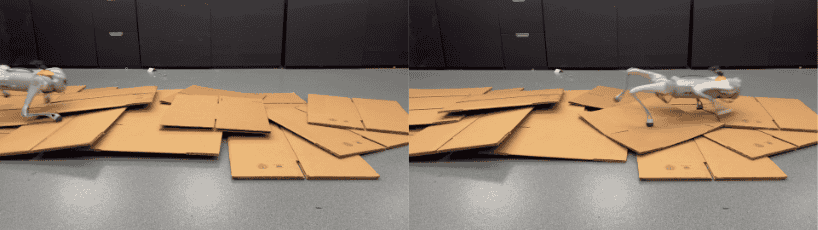}
        \caption{Snapshots of the quadruped robot climbing over randomly distributed paper boards.}
    \end{subfigure}
    \begin{subfigure}{\linewidth}
        \includegraphics[width=\linewidth]{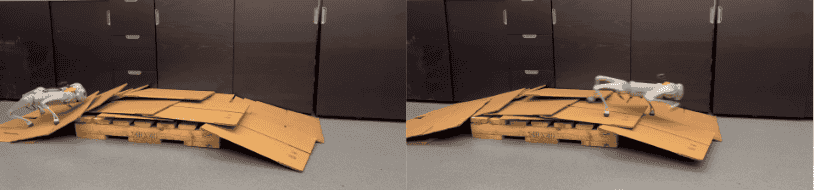}
        \caption{Snapshots of the quadruped robot climbing over a 20 cm step covered by paper boards.}
    \end{subfigure}
\caption{Quadruped robot clearing terrains.}
\label{fig:terrain_snap}
\end{figure}

We tested the trained policy on the real Go1 quadruped robot. Fig. \ref{fig:terrain_snap} shows experiment snapshots. The robot successfully cleared a 20 cm step covered by paper boards, which is a considerable high step compared to the size of Go1.


\section{Discussion and Conclusion}
This paper presented a novel approach to energy-efficient locomotion in quadruped robots by implementing a simplified, energy-centric reward strategy within a reinforcement learning framework. Our method demonstrated that quadruped robots, specifically Unitree Go1, could autonomously develop and transition between various gaits across different linear and angular velocites without relying on predefined gait patterns or intricate reward designs. The adaptive energy reward function, adjusted based on velocity, enabled these robots to select the most energy-efficient locomotion strategies naturally.

There are more extended locomotion tasks that we can make use of the energy-centric training, where the motion reward is no longer velocity tracking but includes more task-specific rewards. This would require the system to dynamically adjust its strategy in response to different task rewards. While this study concentrated on locomotion tasks, the underlying principle of leveraging energy efficiency to drive behavior selection holds broader potential. Future work could explore applying this energy-centric approach across different robotic tasks. For instance, manipulation and interaction tasks could also benefit from strategies prioritizing energy efficiency, potentially finding natural, efficient behaviors analogous to those observed in biological systems \cite{cheng2023legs, vollenweider2023advanced}. Such a framework would align robotic systems more closely with sustainability principles and environmental consciousness.

\section*{Acknowledgements}
This project was supported by the Agency of Science, Technology and Research of Singapore via the National Science Scholarship, the FANUC Advanced Research Laboratory, and the InnoHK of the Government of the Hong Kong Special Administrative Region via the Hong Kong Centre for Logistics Robotics.

\bibliographystyle{IEEEtran}
\bibliography{bibliography.bib}

\begin{thebibliography}{10}
\providecommand{\url}[1]{#1}
\csname url@rmstyle\endcsname
\providecommand{\newblock}{\relax}
\providecommand{\bibinfo}[2]{#2}
\providecommand\BIBentrySTDinterwordspacing{\spaceskip=0pt\relax}
\providecommand\BIBentryALTinterwordstretchfactor{4}
\providecommand\BIBentryALTinterwordspacing{\spaceskip=\fontdimen2\font plus
\BIBentryALTinterwordstretchfactor\fontdimen3\font minus \fontdimen4\font\relax}
\providecommand\BIBforeignlanguage[2]{{%
\expandafter\ifx\csname l@#1\endcsname\relax
\typeout{** WARNING: IEEEtran.bst: No hyphenation pattern has been}%
\typeout{** loaded for the language `#1'. Using the pattern for}%
\typeout{** the default language instead.}%
\else
\language=\csname l@#1\endcsname
\fi
#2}}

\bibitem{cappellini2006motor}
G.~Cappellini, Y.~P. Ivanenko, R.~E. Poppele, and F.~Lacquaniti, ``Motor patterns in human walking and running,'' \emph{Journal of neurophysiology}, vol.~95, no.~6, pp. 3426--3437, 2006.

\bibitem{xi2015selectinggaits}
W.~Xi, Y.~Yesilevskiy, and C.~D. Remy, ``Selecting gaits for economical locomotion of legged robots,'' \emph{The International Journal of Robotics Research}, vol.~35, no.~9, pp. 1140--1154, 2016.

\bibitem{siekmann2021simtoreal}
J.~Siekmann, Y.~Godse, A.~Fern, and J.~Hurst, ``Sim-to-real learning of all common bipedal gaits via periodic reward composition,'' in \emph{IEEE International Conference on Robotics and Automation}, 2021.

\bibitem{margolis2023walk}
G.~B. Margolis and A.~Pulkit, ``Walk these ways: Tuning robot control for generalization with multiplicity of behavior,'' in \emph{Conference on Robot Learning}, 2023.

\bibitem{yang2021fast}
Y.~Yang, T.~Zhang, E.~Coumans, J.~Tan, and B.~Boots, ``Fast and efﬁcient locomotion via learned gait transitions,'' in \emph{Conference on Robot Learning}, 2021.

\bibitem{rudin2021learning}
N.~Rudin, D.~Hoeller, P.~Reist, and M.~Hutter, ``Learning to walk in minutes using massively parallel deep reinforcement learning,'' in \emph{Conference on Robot Learning}, 2021.

\bibitem{feng2023genloco}
G.~Feng, H.~Zhang, Z.~Li, X.~B. Peng, B.~Basireddy, L.~Yue, Z.~Song, L.~Yang, Y.~Liu, K.~Sreenath, \emph{et~al.}, ``Genloco: Generalized locomotion controllers for quadrupedal robots,'' in \emph{Conference on Robot Learning}, 2023.

\bibitem{liang2018gpu}
J.~Liang, V.~Makoviychuk, A.~Handa, N.~Chentanez, M.~Macklin, and D.~Fox, ``Gpu-accelerated robotic simulation for distributed reinforcement learning,'' in \emph{Conference on Robot Learning}, 2018.

\bibitem{makoviychuk2021isaac}
V.~Makoviychuk, L.~Wawrzyniak, Y.~Guo, M.~Lu, K.~Storey, M.~Macklin, D.~Hoeller, N.~Rudin, A.~Allshire, A.~Handa, and G.~State, ``Isaac gym: High performance {GPU} based physics simulation for robot learning,'' in \emph{Conference on Neural Information Processing Systems}, 2021.

\bibitem{mahankali2024maximizing}
S.~Mahankali, C.-C. Lee, G.~B. Margolis, Z.-W. Hong, and P.~Agrawal, ``Maximizing quadruped velocity by minimizing energy,'' in \emph{IEEE International Conference on Robotics and Automation}, 2024.

\bibitem{fu2021minimizing}
Z.~Fu, A.~Kumar, J.~Malik, and D.~Pathak, ``Minimizing energy consumption leads to the emergence of gaits in legged robots,'' in \emph{Conference on Robot Learning}, 2021.

\bibitem{go1}
``Unitree robotics, go1,'' \url{https://www.unitree.com/products/go1}, online; accessed Jun. 2022.

\bibitem{hwangbo2019learning}
J.~Hwangbo, J.~Lee, A.~Dosovitskiy, D.~Bellicoso, V.~Tsounis, V.~Koltun, and M.~Hutter, ``Learning agile and dynamic motor skills for legged robots,'' \emph{Science Robotics}, vol.~4, no.~26, p. eaau5872, 2019.

\bibitem{ji2022concurrent}
G.~Ji, J.~Mun, H.~Kim, and J.~Hwangbo, ``Concurrent training of a control policy and a state estimator for dynamic and robust legged locomotion,'' \emph{IEEE Robotics and Automation Letters}, vol.~7, no.~2, pp. 4630--4637, 2022.

\bibitem{li2021reinforcement}
Z.~Li, X.~Cheng, X.~B. Peng, P.~Abbeel, S.~Levine, G.~Berseth, and K.~Sreenath, ``Reinforcement learning for robust parameterized locomotion control of bipedal robots,'' in \emph{IEEE International Conference on Robotics and Automation}, 2021.

\bibitem{chen2023learning}
S.~Chen, B.~Zhang, M.~W. Mueller, A.~Rai, and K.~Sreenath, ``Learning torque control for quadrupedal locomotion,'' in \emph{IEEE-RAS International Conference on Humanoid Robots}, 2023.

\bibitem{miki2022learning}
T.~Miki, J.~Lee, J.~Hwangbo, L.~Wellhausen, V.~Koltun, and M.~Hutter, ``Learning robust perceptive locomotion for quadrupedal robots in the wild,'' \emph{Science Robotics}, vol.~7, no.~62, p. eabk2822, 2022.

\bibitem{lee2020learning}
J.~Lee, J.~Hwangbo, L.~Wellhausen, V.~Koltun, and M.~Hutter, ``Learning quadrupedal locomotion over challenging terrain,'' \emph{Science Robotics}, vol.~5, no.~47, p. eabc5986, 2020.

\bibitem{choi2023learning}
S.~Choi, G.~Ji, J.~Park, H.~Kim, J.~Mun, J.~H. Lee, and J.~Hwangbo, ``Learning quadrupedal locomotion on deformable terrain,'' \emph{Science Robotics}, vol.~8, no.~74, p. eade2256, 2023.

\bibitem{gabriel2024rapid}
G.~B. Margolis, G.~Yang, K.~Paigwar, T.~Chen, and P.~Agrawal, ``Rapid locomotion via reinforcement learning,'' \emph{The International Journal of Robotics Research}, vol.~43, no.~4, pp. 572--587, 2024.

\bibitem{raibert1990trotting}
M.~H. Raibert, ``Trotting, pacing and bounding by a quadruped robot,'' \emph{Journal of Biomechanics}, vol.~23, pp. 79--98, 1990.

\bibitem{da2020learning}
X.~Da, Z.~Xie, D.~Hoeller, B.~Boots, A.~Anandkumar, Y.~Zhu, B.~Babich, and A.~Garg, ``Learning a contact-adaptive controller for robust, efficient legged locomotion,'' in \emph{Conference on Robot Learning}, 2020.

\bibitem{duan2022sim}
H.~Duan, A.~Malik, J.~Dao, A.~Saxena, K.~Green, J.~Siekmann, A.~Fern, and J.~Hurst, ``Sim-to-real learning of footstep-constrained bipedal dynamic walking,'' in \emph{IEEE International Conference on Robotics and Automation}, 2022.

\bibitem{cleach2024fast}
S.~Le~Cleac'h, T.~A. Howell, S.~Yang, C.-Y. Lee, J.~Zhang, A.~Bishop, M.~Schwager, and Z.~Manchester, ``Fast contact-implicit model predictive control,'' \emph{IEEE Transactions on Robotics}, vol.~40, pp. 1617--1629, 2024.

\bibitem{gabrielli2011price}
G.~Gabrielli, ``What price speed? specific power required for propulsion of vehicles,'' \emph{Mechanical Engineering-CIME}, vol. 133, no.~10, pp. 4--5, 2011.

\bibitem{tucker1975energy}
V.~Tucker, ``The energetic cost of moving about: Walking and running are extremely inefficient forms of locomotion. much greater efficiency is achieved by birds, fish, and bicyclists.'' \emph{American Scientist}, vol.~63, no.~4, pp. 413--419, 1975.

\bibitem{ralston1958speed}
H.~Ralston, ``Energy-speed relation and optimal speed during level walking.'' \emph{Internationale Zeitschrift für angewandte Physiologie einschließlich Arbeitsphysiologie}, vol.~17, no.~4, pp. 277--283, 1958.

\bibitem{umberger2007step}
B.~Umberger and P.~Martin, ``Mechanical power and efficiency of level walking with different stride rates.'' \emph{Journal of Experimental Biology}, vol. 210, no.~18, pp. 3255--3265, 2007.

\bibitem{chevallereau2001bipedal}
C.~Chevallereau and Y.~Aoustin, ``Optimal reference trajectories for walking and running of a biped robot.'' \emph{Robotica}, vol.~19, no.~5, pp. 557--569, 2001.

\bibitem{remy2011bipedal}
C.~D. Remy, ``Optimal exploitation of natural dynamics in legged locomotion.'' Ph.D. dissertation, Eidgenossische Technische Hochschule., 2011.

\bibitem{kiguchi2002quad}
K.~Kiguchi, Y.~Kusumoto, K.~Watanabe, I.~K, and F.~T, ``Energy-optimal gait analysis of quadruped robots.'' \emph{Artificial Life and Robotics.}, vol.~6, no.~3, pp. 120--125, 2002.

\bibitem{muraro2003quad}
A.~Muraro, C.~Chevallereau, and Y.~Aoustin, ``Optimal trajectories for a quadruped robot with trot, amble and curvet gaits for two energetic criteria.'' \emph{Multibody System Dynamics}, vol.~9, no.~1, pp. 39--62, 2003.

\bibitem{zhuang2023robot}
Z.~Zhuang, Z.~Fu, J.~Wang, C.~Atkeson, S.~Schwertfeger, C.~Finn, and D.~Pathak, ``Robot parkour learning,'' in \emph{Conference on Robot Learning}, 2023.

\bibitem{cheng2023legs}
X.~Cheng, A.~Kumar, and D.~Pathak, ``Legs as manipulator: Pushing quadrupedal agility beyond locomotion,'' in \emph{IEEE International Conference on Robotics and Automation}, 2023.

\bibitem{vollenweider2023advanced}
E.~Vollenweider, M.~Bjelonic, V.~Klemm, N.~Rudin, J.~Lee, and M.~Hutter, ``Advanced skills through multiple adversarial motion priors in reinforcement learning,'' in \emph{IEEE International Conference on Robotics and Automation}, 2023.

\end{thebibliography}

\end{document}